\documentclass[10pt,twocolumn,letterpaper]{article}

\usepackage{cvpr}              % To produce the CAMERA-READY version
\usepackage[dvipsnames]{xcolor}

\usepackage{bbm}
\usepackage{multirow}
\usepackage{booktabs}
\usepackage{subcaption}
\usepackage{booktabs}
\usepackage{colortbl}
\usepackage{booktabs} 
\usepackage{amssymb}% http://ctan.org/pkg/amssymb
\usepackage{pifont}% http://ctan.org/pkg/pifont
\usepackage{bm}
\usepackage[accsupp]{axessibility}

\newcommand{\cmark}{\ding{51}}%
\newcommand{\xmark}{\ding{55}}%
\definecolor{mygray}{gray}{.9}

\newcommand{\bq}{\bm{q}} % query image
\newcommand{\bv}{\bm{v}} % multi-view images
\newcommand{\bx}{\bm{x}} % grid of 3d coordinates, 3 x h x w
\newcommand{\bz}{\bm{z}}
\newcommand{\bff}{\bm{f}}
\newcommand{\bF}{\bm{F}}
\newcommand{\bM}{\bm{M}}
\newcommand{\bP}{\bm{P}}
\newcommand{\bp}{\bm{p}}
\newcommand{\mcV}{\mathcal{V}}

\makeatletter
\DeclareRobustCommand\onedot{\futurelet\@let@token\@onedot}
\def\@onedot{\ifx\@let@token.\else.\null\fi\xspace}
\def\eg{\emph{e.g}\onedot,\xspace}

\def\wrt{w.r.t\onedot} 
\def\etal{\emph{et al}\onedot}
\renewcommand{\paragraph}{%
	\@startsection{paragraph}{4}{\z@}%
	{0.1em \@plus 0.5ex \@minus 0.2ex}{-1em}%
	{\normalsize\bf}%
}
\makeatother

\definecolor{cvprblue}{rgb}{0.21,0.49,0.74}
\usepackage[pagebackref,breaklinks,colorlinks,citecolor=cvprblue]{hyperref}

\newcommand{\dname}{\emph{BrokenChairs-180K}\xspace}
\newcommand{\ourmethodlong}{\emph{Correspondence Matching Transformer}\xspace}
\newcommand{\ourmethod}{\emph{CMT}\xspace}
\usepackage[capitalize]{cleveref}
\crefname{section}{Sec.}{Secs.}
\Crefname{section}{Section}{Sections}
\Crefname{table}{Table}{Tables}
\crefname{table}{Tab.}{Tabs.}

\def\paperID{7563} % *** Enter the Paper ID here
\def\confName{CVPR}
\def\confYear{2024}

\title{Looking 3D:~Anomaly Detection with 2D-3D Alignment}
\author{Ankan Bhunia \quad Changjian Li \quad Hakan Bilen \\
University of Edinburgh \vspace{1mm}\\
\href{https://groups.inf.ed.ac.uk/vico/research/Looking3D}{https://groups.inf.ed.ac.uk/vico/research/Looking3D}
}
\begin{document}
\maketitle
\begin{abstract}
% Automatic anomaly detection from visual cues is of significance in practical applications, e.g., furniture manufacturing and product quality measurement. 
% In this paper, we, for the first time, introduce a new reference-based anomaly detection problem, i.e., identify anomalies in a query image by comparing it to a reference shape. 
% To solve it, we have synthetically built a large dataset (i.e., \dname) comprising 180$K$ images, with over 100$K$ of them featuring some form of anomaly, 
% and proposed a novel deep learning approach, which learns the correspondence between the query image and the reference 3D shape and detects anomalies using the attention mechanism.  
% We have conducted comprehensive experiments to validate our superior performance, which will serve as the benchmark for future research.
Automatic anomaly detection based on visual cues holds practical significance in various domains, such as manufacturing and product quality assessment. 
This paper introduces a new conditional anomaly detection problem, which involves identifying anomalies in a query image by comparing it to a reference shape. 
To address this challenge, we have created a large dataset, \dname, consisting of around 180$K$ images, with diverse anomalies, geometries, and textures paired with 8,143 reference 3D shapes. 
To tackle this task, we have proposed a novel transformer-based approach that explicitly learns the correspondence between the query image and reference 3D shape via feature alignment and leverages a customized attention mechanism for anomaly detection. 
Our approach has been rigorously evaluated through comprehensive experiments, serving as a benchmark for future research in this domain. 
% The dataset and code are available at: \href{https://github.com/VICO-UoE/Looking3D}{https://github.com/VICO-UoE/Looking3D}.
% The dataset will be made publicly available. 
\end{abstract}    
\section{Introduction}
\label{sec:intro}

Anomaly detection (AD)~\cite{chandola2009anomaly,pang2021deep}, identifying instances that are irregular or significantly deviate from the normality, is an actively studied problem in several fields.
In standard vision AD benchmarks, `irregularities' are typically caused by either high-level (or semantic) variations such as presence of objects from unseen categories \cite{ahmed2020detecting,blum2021fishyscapes,chan2021segmentmeifyoucan}, defects such as scratches, dents on objects \cite{bergmann2021mvtec}, low-level variations in color, shape, size \cite{deecke2021transfer}, or pixel-level noise \cite{hendrycks2016baseline}.
The standard approach has been to learn representations along with classifiers that are robust to the variations within the regular set of instances, and, at the same time, sensitive to the ones causing irregularities.
However, this paradigm performs poorly when the irregularities are arbitrary and conditional to the context and/or individual characteristics of the instance which may not be known in prior or observed.
For instance, in an object category such as `chair' that contains visually very diverse instances with huge intra-class variation, having three legs may imply a missing leg and hence an anomaly for a chair instance, while regularity for another instance.
The AD here depends on whether the chair instance was originally designed to have three legs.
\begin{figure}[t!]
\begin{center}
   \includegraphics[width=0.99\linewidth]{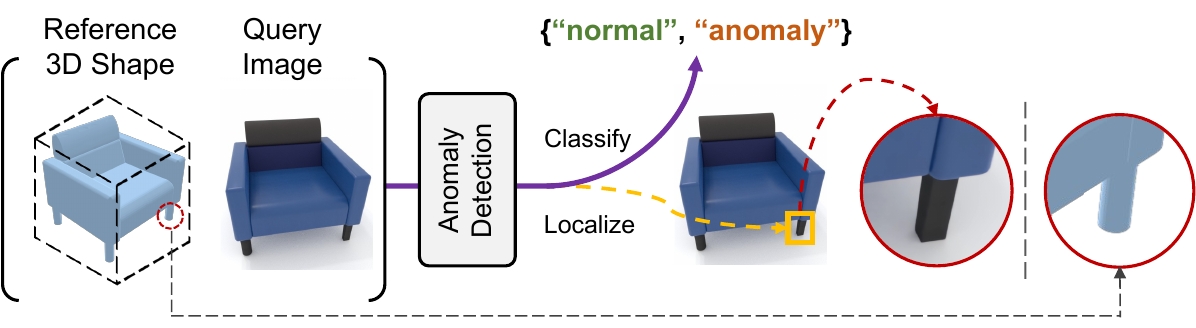}
\end{center}
\vspace{-0.6cm}
\caption{We propose a new conditional AD task that aims to identify and localize anomalies in a query image by comparing it to a reference shape. 
% If the query image contains an anomaly then we identify it and optionally obtain a bounding box locating the anomalous region. 
The anomalous region is shown in a yellow bounding box. 
For instance, the right leg of the blue sofa is rectangular unlike the cylindrical one in its reference shape. 
%% zoomed regions contain noise. need to try rendering at a higher resolution for these two?
% \CJ{Check: the chair 3D shape in the first row is not exactly the same as in the rendered RGB image, they are different in terms of armrest and back. Update: the second example in the second row does not show a similar view of the query image, which does not help the reader to check if our prediction is correct or not. But this case can be decided solely in the image, it is a bit distracting. Edit: put 'Localization' around the dashed arrow and put 'Identification' to the solid arrow, meaning we target two tasks: identification/classification and localization, which is consistent with the caption. Minor: I prefer the figure without the outer dashed box.}
}

\label{fig:intro}
 \vspace{-0.4cm}
\end{figure}

Motivated by the intuition above, this paper introduces a novel \emph{conditional} AD task, along with a new benchmark and an effective solution,  
that aims to identify and localize anomalies from a photo of an object instance (i.e., the \emph{query image}), in relation to a reference 3D model (see \cref{fig:intro}).
The 3D model provides the reference shape for the regular object instance, and hence a clear definition of regularity for the query image.
This setting is motivated by real-world applications in inspection and quality control, where an object instance is manufactured based on a reference 3D model, which can then be used to identify anomalies (\eg production faults, damages) from a photo of the instance.

The proposed task goes beyond the single image analysis in standard AD benchmarks and requires the detection of subtle anomalies in shape by comparing two modalities, an image with its reference 3D model, which is challenging for three reasons.
First, we would like our model to detect anomalies in previously unseen object instances from image-shape pairs at test time.
Generalizing to unseen instances demands learning rich representations encoding a diverse set of 3D shapes and appearances while enabling accurate localization of anomalies.
Second, the reference 3D model contains only shape but not texture information to simulate a realistic scenario where the 3D model can be used to produce instances with different materials, colors, and textures.
The resulting domain gap between two modalities requires learning representations that are invariant to such appearance changes and sensitive to variations in geometry.
Finally, in our benchmark, the viewpoint of the object instances in query images is not available in training.
This requires the model to establish the local correspondences between the modalities, \textit{i.e.}, corresponding 3D location for each image patch in an unsupervised manner.

To tackle the first challenge, we propose a new large-scale dataset, \dname, consisting of around 180$K$ query images with diverse anomalies, geometries, and textures paired with 8,143 reference 3D shapes.
Training on such a diverse dataset enables learning rich multi-modal representations to generalize to unseen objects.
To address the domain gap between the query images and reference shapes, we follow two strategies.
First, we render each reference shape from multiple viewpoints to generate a set of multi-view images to represent the 3D shape and use them as input along with a query image to our model.
The multi-view representation facilitates learning domain-invariant representations through sharing the same encoder across query and multi-view images.
Second, our model, \ourmethodlong (\ourmethod) learns to capture cross-modality relationships by applying a novel cross-attention mechanism through a sparse set of local correspondences.
Finally, to address the third challenge, we use an auxiliary task that forces the model to learn viewpoint invariant representations for each local patch in query and multi-view images enabling our method to align local features corresponding to the same 3D location regardless of its viewpoint {without ground truth correspondences}.
% Finally, we use an auxiliary task, along with the AD task, that forces the model to learn viewpoint invariant representations for each local patch in query and multi-view images enabling our method to align local features corresponding to the same 3D location regardless of its viewpoint {without groundtruth correspondences}.

% Most related to our method, image-based 3D shape retrieval techniques~\cite{grabner20183d,grabner2019location,lin2021single}  aim to retrieve the most similar shape in a database for a given 2D image. 
% While these methods involve cross 2D-3D modeling by matching their global embeddings, our task crucially requires capturing finer-grained local details for which we use a dense correspondence matching technique.
% Compared to existing image-based anomaly techniques~\cite{}, our method differs in its focus, detects anomalies in local shape from images by comparing to a 3D reference model.

In summary, our main contributions are threefold, introducing a novel AD task, a large-scale benchmark to provide a testbed for future research, and a customized solution.
Our model includes multiple technical innovations including a hybrid 2D-3D representation for 3D shapes, a transformer-based architecture that jointly learns to densely align query and multi-view images from image-level supervision and detect anomalies.
Our results in extensive ablation studies clearly demonstrate that 3D information along with correspondence matching yields significant improvements.
We also perform an additional perceptual study that evaluates the human performance on the task, showing that the proposed task is challenging.
Finally, we evaluate our technique on real images showing promising results.

\section{Related Work}
\label{sec:related_work}

\noindent\textbf{AD methods.} We refer to \cite{chandola2009anomaly,pang2021deep} for detailed literature reviews.
Unlike the standard AD techniques, we focus on a conditional and multi-modal AD problem which requires a joint analysis of a query image with a reference 3D shape to detect local irregularities in the image.

% AD has been extensively studied across diverse fields. 
% While well-known classical methods use Bayesian networks~\cite{manikopoulos2002network}, rule-based systems~\cite{pang2019deep}, and clustering algorithms~\cite{sequeira2002admit}, many recent ones leverages powerful deep neural networks. 
% Most methods are designed as unsupervised learning due to the lack of large-scale labeled anomaly data. 
% Among these, reconstruction-based methods~\cite{chen2017outlier, zhou2017anomaly, mishra2020neural, liu2020towards, sabokrou2018adversarially, pidhorskyi2018generative} involve training auto-encoders~\cite{mishra2020neural,liu2020towards} or GAN-based networks~\cite{sabokrou2018adversarially, pidhorskyi2018generative} to reconstruct the input image. When trained only on normal data, they are expected to poorly reconstruct anomalies, with the reconstruction error serving as the anomaly score. 
% To address limited labeled data, by using GANs~\cite{pourreza2021g2d, zaheer2020old, murase2022algan, chatillon2021history} or diffusion models~\cite{mirzaei2022fake}, some recent methods generate pseudo-anomalous data which is further used to train supervised AD methods.
% Unlike these methods, we focus on a conditional AD problem which requires a joint analysis of a query image with a 3D reference model to detect local irregularities in shape.

\begin{figure*}[t!]
\begin{center}   \includegraphics[width=1\linewidth]{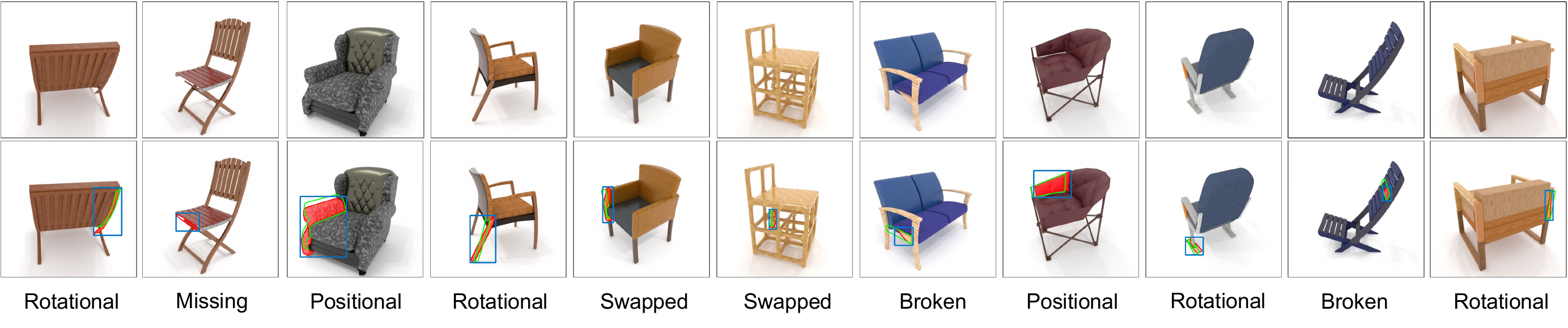}
\end{center}
\vspace{-7mm}
\caption{\textbf{Example anomaly instances from our \dname dataset.} Our dataset consists of around 100$K$ anomaly images. In the top row, some example anomaly instances are shown, along with the ground truth bounding boxes and segmentation masks in the bottom row. The red mask is used to indicate parts with anomalies, and a green contour line highlights their respective regions prior to applying any anomaly, and the bounding box is shown as blue rectangular boxes. \textit{(figure best viewed in zoom)} 
% \CJ{Update: remove (a), (b) to save space for the figures. The second, the seventh, and the second-last examples should rotate to the best view to see the anomaly clearly. Move the anomaly type to the bottom-right of the blue box, since it will occlude important information.}
% \CJ{any transformations - applying any designed anomaly}
}
\label{fig:data}
 \vspace{-5mm}
\end{figure*}

\noindent\textbf{Conditional/referential AD.}
In many AD applications, the anomaly of an instance depends on its specific context~\cite{song2007conditional}.
For instance, anomalous temperature changes can be more accurately detected in a particular spatial and temporal context.
We also study a specific application of the conditional AD problem where the context information is instance-specific and comes from a reference 3D shape.
% This referential AD problem is especially common in scenarios where variations in the set of regular instances are significant and difficult to separate them from factors causing anomalies.

\noindent\textbf{AD image benchmarks.} 
A major problem in the development of AD is the lack of large datasets with realistic anomalies.
For semantic anomalies, a common practice (\eg~\cite{chalapathy2018anomaly, ruff2018deep}) is to select an arbitrary subset of classes from an existing classification dataset (\eg MNIST~\cite{lecun1998gradient}, CIFAR10~\cite{krizhevsky2009learning}), treat them as an anomalous class, and train a model only on the remaining classes. 
% While this strategy enables obtaining a large training set, the anomalous instances are usually significantly different from the regular ones. 
% Hence, it remains unclear how these models perform when irregularities are caused by subtler factors.
There also exist multiple datasets that contain real-world anomalous instances including irregularly shaped objects~\cite{saleh2013object}, objects with various defects such as scratches, dents, contaminations~\cite{bergmann2019mvtec}, various defects in nanofibrous material~\cite{carrera2016defect} which focus on one sample at a time.
% Different to the existing benchmarks focusing on one sample at a time, we propose a multi-modal benchmark for conditional AD.
{
% Two concurrent work, Real3D-AD~\cite{liu2024real3d} and PAD~\cite{zhou2024pad} respectively focusing on AD from only 3D input 
A concurrent work, PAD~\cite{zhou2024pad} targets a similar objective with ours, while our task has fewer assumptions and is designed to detect fine-grained geometrical anomalies. 
Moreover, compared to the PAD dataset consisting of only 20 LEGO bricks of animal toys, ours comprises a large-scale collection of realistic chairs with diverse geometries, textures, and a wider range of fine-grained anomalies.}

\noindent\textbf{2D-3D cross-modal correlation.} Image-based 3D shape retrieval~\cite{grabner20183d,grabner2019location,lin2021single} is a related problem that aims to retrieve the most similar shape for a given 2D image. 
Most existing works learn to embed 2D images and 3D shapes into a common feature space and perform metric learning using a triplet loss. 
Different to the retrieval task that primarily involves global-level matching, our focus is comprehending the correlation of fine-grained local details between the shape and the image to detect anomalies within the image. 
% Aubry~\etal~\cite{aubry2014seeing} propose detecting object instances in images by aligning them with the corresponding 3D shape through exemplar part based representations.
% Aubry~\etal~\cite{aubry2014seeing} use a novel 2D-to-3D alignment problem that not only retrieves a 3D model matching the style of the 2D image but also recovers its viewpoint relative to the camera, resulting in an aligned 3D view with the 2D image. 
Another related area focuses on learning of 2D-3D correspondences~\cite{feng20192d3d, pham2020lcd, wang2023dgc, li20232d3d} by matching 2D and 3D locally with a triplet loss~\cite{feng20192d3d, pham2020lcd}, matching images and point clouds with a coarse-to-fine approach~\cite{li20232d3d},  improving matching robustness using  a global-to-local
Graph Neural Network~\cite{wang2023dgc}.
2D-3D correlation is also studied for specific applications such as object pose estimation~\cite{lim2013parsing, xiao2012localizing}, 3D shape estimation~\cite{hejrati2012analyzing} and object detection in images by using a set of 3D models~\cite{aubry2014seeing}.  
Unlike the methods discussed here, our objective is to identify and localize anomalies in a given 2D query image in relation to a reference 3D model. 

\section{Building \dname Dataset}
\label{sec:data}
To the best of our knowledge, there is no prior large public dataset with paired 3D shapes and images. Hence we introduce \dname, a new benchmark for the proposed conditional AD task.
% Here we introduce a new benchmark for the proposed AD task, describe the procedure used in its generation and provide various details.
% To the best of our knowledge, there is no prior public dataset focusing on anomaly detection from shape and image pairs.
% Since there is no large-scale public dataset containing labeled anomaly data, in this section, we describe our framework for building a synthetic dataset for reference-based anomaly detection tasks.
Our dataset focuses on generating samples from one category, namely `chair', which includes various subcategories like sofas, office chairs, and stools, while our generation pipeline is general and applicable to other categories.
We picked this category as chairs contain a very wide range of shapes, appearances, and material combinations making them appealing for our experiments.
%In addition, 3D shapes of chairs can be found in prior 3D datasets such as the PartNet~\cite{mo2019partnet} which can be costly to obtain otherwise. 
In the following, we describe the generation procedure, including anomaly creation and realistic image rendering. More details can be found in the supplementary.

% In this work, our main focus has been on chairs, which include various types like sofas, office chairs, stools, and more. Chairs have a diverse set of appearances and material combinations that make them appealing for our experiments, however, our dataset creation pipeline also applies to other categories. %Next, we describe the pipeline starting from 3D shape collections, then applying various forms of geometric deformations to the shapes to create anomalies, followed by rendering those shapes with photorealistic material, and finally a quality control step to verify the anomalies.

\subsection{Creating Anomaly from 3D Objects}
\label{subsec:create_anomaly}
\noindent \textbf{3D shape collection.} To cover a wide variety of fine-grained anomalies across various parts of chairs (\eg leg, arm, and headrest), we strive to collect 3D shapes that come with part annotations and thus opt to utilize the PartNet~\cite{mo2019partnet} as our starting point. 
PartNet is a large-scale dataset of 3D objects annotated with fine-grained part labels. 
Its chair category is among the most populous, providing a rich source of 3D shapes for our task. 
In particular, we use 8,143 3D chair shapes from PartNet.
Given a 3D model of a chair and its part annotation, we automatically create anomalies by applying geometric deformations as described below.

\noindent \textbf{Generation of anomaly shapes.} Our dataset covers five anomaly scenarios (see \cref{fig:data}) relevant to real-world applications. 
% These scenarios can be broadly categorized into five attributes: %\textit{positional anomalies}, \textit{rotational anomalies}, \textit{damaged or broken parts}, \textit{component swapping}, and \textit{missing components}.  
\noindent \textit{(1) Positional anomalies} pertain to deviations from the designated position of chair parts. %For example, if the legs of a chair are not evenly aligned, it can lead to stability issues.
To create a positional anomaly, we randomly select a part from a normal 3D model and apply random translation. % of $\delta p$ which is randomly sampled for each axis from a uniform distribution of range $0.04\leq |\delta p| \leq 0.08$. 
\noindent \textit{(2) Rotational anomalies} are created by applying a 3D rotational transformation to a randomly selected 3D part. 
%The rotation matrix is formed using a random rotation axis and an angle $r_\theta$ (radians) sampled from a uniform distribution of range $0.2\leq |r_\theta| \leq 0.4$. The center of rotation is set to a fixed point at one of the connecting points between the anomalous part and the main body of the chair object. 
\noindent \textit{(3) Broken or damaged parts} consist cases that structural components are broken or damaged.
We synthetically generate breaks using Boolean subtraction following~\cite{lamb2022deepjoin} where we fracture a chair part by subtracting a random spherical or cubical geometric primitive from the part mesh. 
%If a break removes more than 90\% or less than 10\% of an object we discard the sample and regenerate another break. 
\noindent \textit{(4) Component swapping} involves swapping common parts across different chair instances (\eg `back-connector' of one chair is exchanged with a `back-connector' from another chair), simulating an incorrect assembly during manufacturing. 
% For example, if we replace one leg of an instance with a leg from a different instance, it can lead to an unstable chair that wobbles or does not sit level. 
% We implement part swapping within identical categories. For instance, a part categorized as a `back-connector' will be exchanged with a `back-connector' from a different chair. 
\noindent \textit{(5) Missing components} involve randomly choosing one part and removing it from the 3D shape. 
Next, we discuss the generation of query images with photo-realistic texture. %\CJ{Further compress if needed.}

% \begin{figure}[t!]
% \begin{center}
%    \includegraphics[width=1\linewidth]{figs/part_stats.pdf}\vspace{-0.5cm}
% \end{center}\vspace{-0.2cm}
% \caption{\textbf{Distribution of anomaly types within our dataset, categorized by salient chair-parts.} Each row in the plot represents the number of instances, various anomaly types are observed within the referenced part in the whole dataset. }
% \label{fig:partstats}
%  \vspace{-0.3cm}
% \end{figure}

\begin{table}[t!]
\begin{center}
\caption{\textbf{Dataset statistics}. The first row shows the number of distinct 3D chair instances utilized in our dataset, the following row indicates the total number of images rendered from these shapes, and the third row denotes [min., max., median] values corresponding to the number of views rendered for each shape object. %Finally, the last row shows  the number of distinct textures that were utilized in the rendering of these chair objects.
% \CJ{Why normal images are rendered from max=12 views? Just to get more images?}
}\vspace{-0.3cm}
\label{tab:stats}
\setlength{\tabcolsep}{2.8pt}
\scalebox{0.7}{
\begin{tabular}{l|c|c|c|c|c|c|c} 
\toprule
\rowcolor{mygray} & \multicolumn{5}{c|}{Different types of Anomalies} & & \\
\rowcolor{mygray}
 & Position & Rotation & Broken & Swapped  & Missing  & \multirow{-2}{*}{\cellcolor{mygray} \#Anomaly } & \multirow{-2}{*}{\cellcolor{mygray} \#Normal } \\ \midrule
\textbf{\#Shapes}      &5,646 &5,551 &6,113 &6,427 &5,182 &28,919  &8,143\\
\textbf{\#Images}  &20,023 &20,017 &20,008 &20,010 &20,023 &100,076 &77,994\\  
\textbf{\#Views}     & [1,7,4]  & [1,7,4] & [1,5,4] & [1,7,4]  & [1,8,4] & [1,8,4] & [1,12,8]\\
\bottomrule
\end{tabular}
}
\end{center} \vspace{-0.8cm}
\end{table}

\subsection{Photo-realistic Rendering of 3D objects}

\noindent \textbf{Assigning materials to 3D shapes.} The shapes in PartNet only contain basic textures but no realistic materials. To enable realistic rendering, we use photo-realistic relightable materials from \cite{park2018photoshape} represented as SVBRDF. 
In total, we utilize 400 publicly available SVBRDF materials, encompassing various types such as wood, plastic, leather, fabric, and metal. 
Following PhotoShape~\cite{park2018photoshape}, we automatically assign a material to each semantic part of a 3D shape, and use Blender's ``Smart UV projection'' algorithm to estimate the UV maps needed for texturing. 

%\noindent \textbf{Representing 3D shape.}

\noindent \textbf{Rendering and view selection.} We render each shape from various viewpoints sampled from a hemisphere around the object. The viewpoint is parameterized in spherical coordinates where azimuth values are sampled uniformly over $\left[0, 2\pi\right)$ with an interval of $\pi/10$ and elevation values are uniformly sampled in $[\pi/9,2\pi/9]$. The radius is fixed at $2.5$ for all views. 
For anomaly shapes, we only keep the rendering if the anomalous part is visible from the camera view. 
We employ a quality control and verification step (see supplementary) to discard bad-quality samples. 

%\noindent \textbf{Quality control and verification. } We aim to ensure that the resulting anomaly shapes adhere to the principles of physics. For instance, if a part detaches from the chair's main body (such that it floats in space) during the deformation process, we reject the generated anomaly and initiate the process again, adjusting the parameters as necessary. See supplementary for more details.
%After 5 attempts we discard the specific part and try with a different part. We also apply an IoU thresholding technique (refer to Supplementary for detailed discussion) to filter out rendered samples in cases where the created anomaly closely resembles its normal counterpart primarily because of the camera viewpoint. To do this, we generate masks for the working part both before and after applying the deformation. Subsequently, we calculate the IoU between these two masks, and if the IoU exceeds 0.8, we discard the rendered image and try with a different camera angle. If the same occurs four times in a row, we will discard that particular anomaly shape.

\begin{figure*}[t!]
\begin{center}
   \includegraphics[width=1\linewidth]{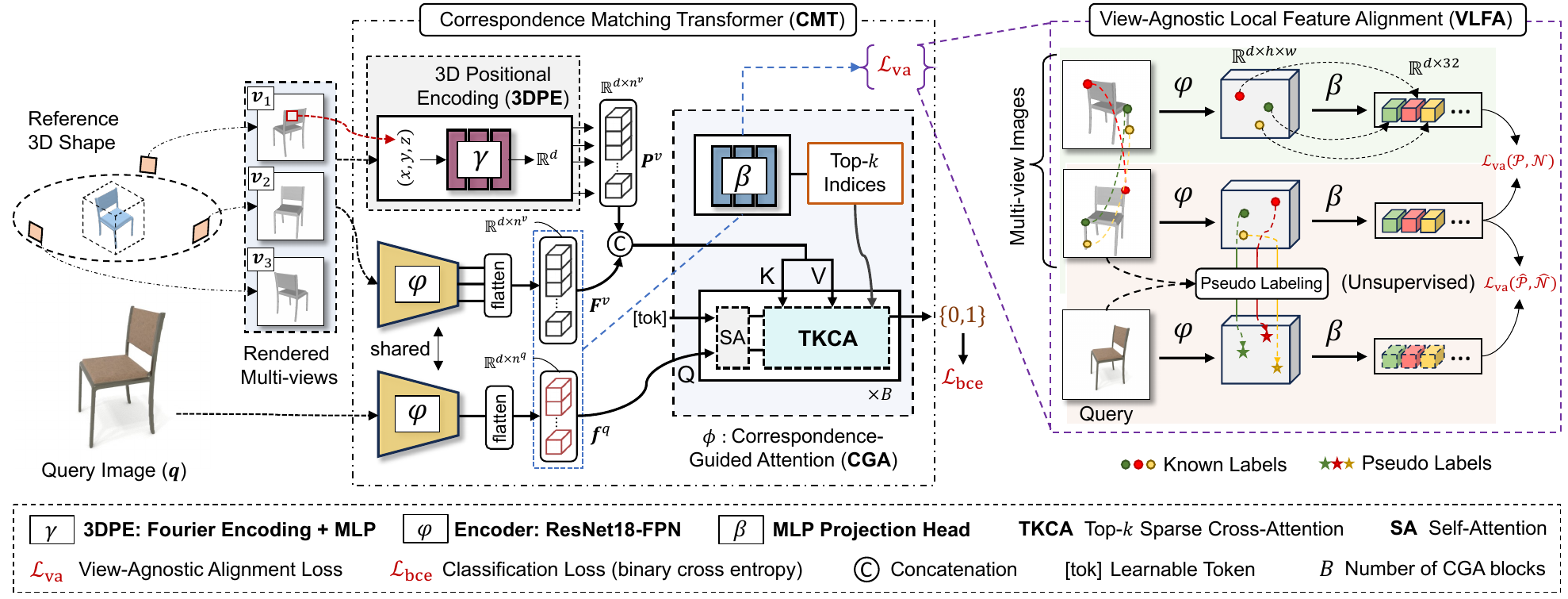}
\end{center}
\vspace{-0.6cm}
\caption{Overall architecture of our proposed CMT framework for conditional AD task. Our CMT takes the following inputs: the query image $\bq$ and the rendered multi-view images $\{\bv_{n}\}_{n=1}^N$. We extract query features $\bff^q$ and multi-view features $\bF^v$ using the encoder $\varphi$. Additionally, we use \textit{3D positional encoding} (3DPE) to obtain 3D positional features $\bP^v$ for the multi-view images. Next, $\bF^v$ and $\bP^v$ are concatenated and fed to the \textit{correspondence-guided attention} (CGA) network, denoted as $\phi$, along with the query features $\bff^q$. The CGA network selectively conditions the final prediction on a small subset of the most related patches from multi-view images through a \textit{top-$k$ sparse cross-attention} (TKCA) mechanism. The \textit{view-agnostic local feature alignment} (VLFA) serves to align the encoder output features to achieve view-agnostic representation through semi-supervised learning. %The view-agnostic features obtained are utilized to establish dense correspondences between similar regions in the query and the multi-view images. These learned correspondences, in turn, guide the CGA network.
% \CJ{Update: $L^{cl}$ uses different fonts in the legend and in the figure.}
}  

\label{fig:method}
 \vspace{-0.5cm}
\end{figure*}

%For instance, when rendering from a particular viewpoint, rotated parts may appear identical to their pre-rotation state, especially when the camera direction aligns with the rotation axis.
\noindent \textbf{Dataset Statistics.} 
% what we have in the dataset
Our dataset comprises a total of 8,143 reference 3D shapes (normal), along with around 180$K$ images rendered at a resolution of $256\times256$ pixels. Among these images, 100$K$ contains anomalies, while the remaining are categorized as normal. 
% how we use the 3D shape 
Since in our solution, we use textureless multi-view images to represent the reference 3D shape, we further provide grayscale multi-view images\footnote{In practice, we clone grayscale values and convert each view to a three-channel image before feeding them as input to our model.} rendered from 20 regularly sampled viewpoints for each reference shape. However, the 3D representation is not necessary to be multi-view images, alternative representations like mesh, point cloud, or voxel can be obtained from the reference shape and adopted by future algorithms when solving the conditional AD problem.

% statistics and splitting
A detailed breakdown of these statistics is provided in \cref{tab:stats}. We divided the dataset into three distinct sets: 138$K$ images for training, 13$K$ for validation, and 26$K$ for testing. Each set contains rendered images from a set of mutually exclusive 3D shapes. 
Hence, the evaluation is performed on \textit{previously unseen} 3D shapes.
%Our dataset has a total of 103 distinct part categories that involve some type of anomaly. The distribution of these parts is illustrated in Fig.~\ref{fig:partstats}. For the visualization, certain part categories have been grouped and are shown in the figure. 
Our dataset also contains bounding box and segmentation mask, localizing any anomalous region. 
%Our dataset is accompanied by a variety of annotations, including a 2D bounding box that encloses any anomalous regions within the rendered images and a 9-DOF bounding box for each reference shape, indicating the anomaly's location relative to a rendered image. 
%Each anomaly image also includes a binary mask locating the anomaly. 
%Fig.~\ref{fig:genstats} shows statistics of bounding box sizes (left) and a plot illustrating the degree of occlusion within our dataset (right). 

\section{Proposed Method}
\label{sec:method}

\subsection{Overview}

Let $\bq \in \mathbb{R}^{3 \times H \times W}$ be an $H\times W$ dimensional RGB image of an object captured from an unknown viewpoint and $\mcV=\{\bv_n\}_{n=1}^N$ be a set of $H\times W$ dimensional images that are rendered from the reference shape at $N$ regularly sampled viewpoints on a hemisphere. {We assume the model has access to the camera pose and depth map of each multi-view image.}
We wish to learn a classifier $\psi: \mathbb{R}^{3 \times H \times W} \times \mathbb{R}^{N \times 3 \times H \times W} \rightarrow [0,1]$ that takes in $\bq$ and $\mcV$ and predicts the ground-truth binary anomaly label $y \in \{0,1\}$.
%  and bounding box $\bb \in \mathbb{R}^4$.
Given a labeled training set $\mathcal{D}$ including $|\mathcal{D}|$ query, multi-view, and label triplets $(\bq,\mcV,y)$, the classifier can be optimized by minimizing the loss term:
\begin{equation}
   \mathcal{L}_\text{bce}(\mathcal{D}) = \sum_{(\bq,\mcV,y)\in \mathcal{D}} \ell_\text{bce}(\psi(\bq,\mcV),y)
   \label{eq:optceonly}
\end{equation} where $\ell_\text{bce}$ is the binary cross-entropy loss function.

% This is challenging for three main reasons.
% First, the correspondences between the query and multi-view images are not known due to absence of ground-truth viewpoint for the query image.
% Second, even when the closest views are found, their viewpoints will not be exactly the same with the viewpoint of the query image, hence demanding certain understanding of 3D to interpolate between the views.
% Third, the multi-view images encode only the shape information due to the unknown texture, lighting and color and hence look different from the query image.

An ideal classifier $\psi$ must identify subtle shape irregularities in $\bq$ by finding the relevant patches in $\mathcal{V}$ for each patch in $\bq$ and comparing them.
One straightforward design to relate patches across query and multi-view images is to use the cross-attention module~\cite{vaswani2017attention}.
In particular, one can use local features extracted from $\bq$ as query and ones from $\mcV$ as key and value matrices as input to the scaled dot-product attention in \cite{vaswani2017attention} to cross-correlate them while predicting the anomaly label.
While this design can implicitly capture such cross-correlations between patches from only image-level supervision when trained with the loss in \cref{eq:optceonly}, it fails to perform better than a similar model that is trained \emph{only} on the query images in practice (see \cref{sec:exp}). 
We posit that the failure to utilize $\mcV$ is due to the difficulty in establishing the correct correspondences from noisy correlations between all patches pairs across query and multi-view images
% (\ie $h \times w \times N \times h \times w $) 
only from image-level supervision. 
% \AB{h,w not defined}

To address this challenge, we propose a new model, \textit{correspondence matching transformer} (CMT) that consists of a CNN encoder, a \textit{3D positional encoding} (3DPE) module, a \textit{correspondence-guided attention} (CGA) network, and lastly a \textit{view-agnostic local feature alignment} (VLFA) mechanism (see \cref{fig:method}).
While the 3DPE module encodes the 3D location of the patches in multi-view images and facilitates finding local correspondences across views, the CGA network selectively conditions the final prediction on a small subset of the most related patches from multi-view images through a top-\textit{k} sparse cross-attention (TKCA) mechanism.
Finally, VLFA provides a richer supervision signal to establish correspondences between similar regions in the query and the multi-view images by using semi-supervised learning. 
Next, we describe them in detail.

\begin{figure}[t!]
\begin{center}
   \includegraphics[width=1\linewidth]{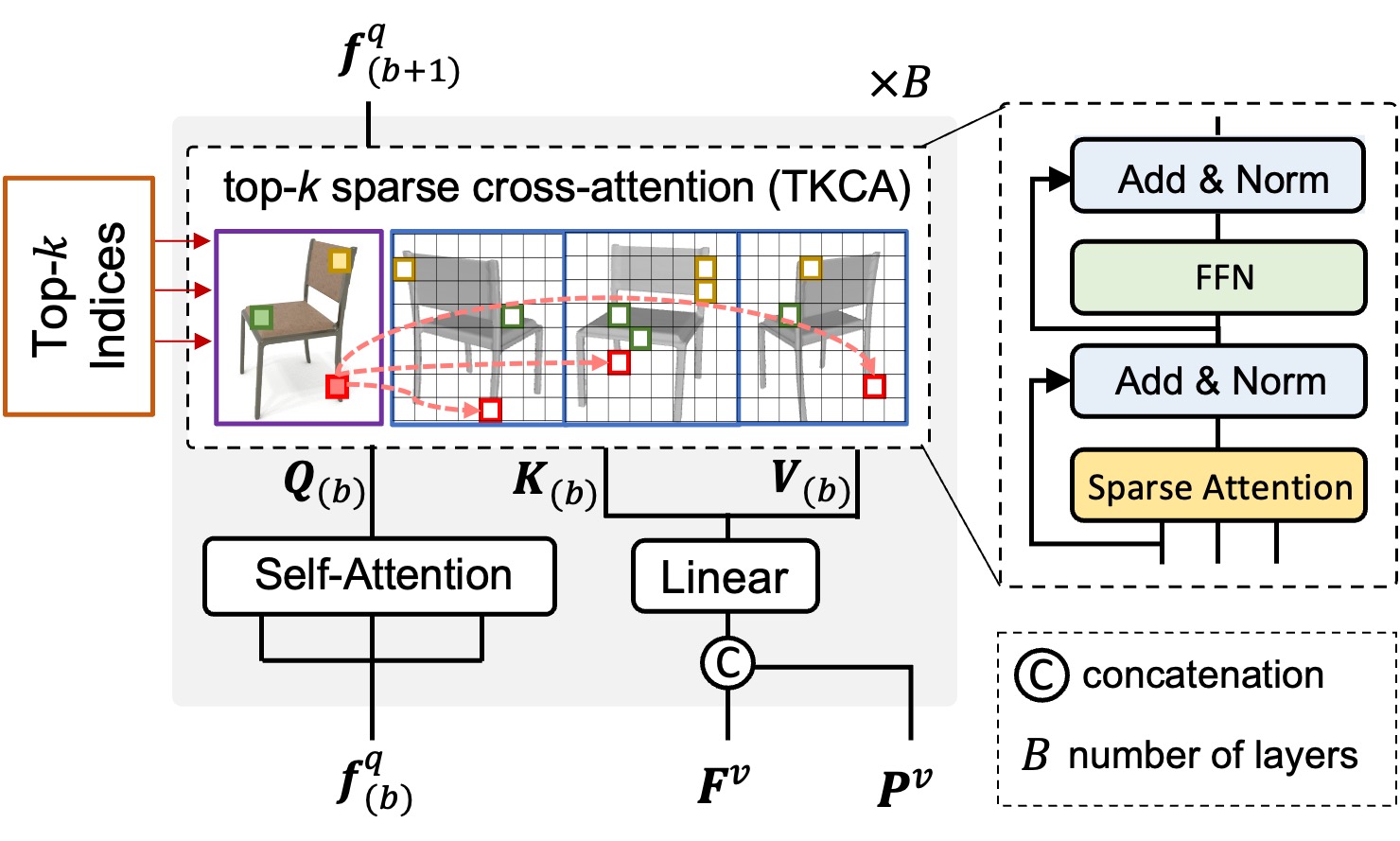}\vspace{-0.5cm}
\end{center}\vspace{-0.3cm}
\caption{ Our proposed \textit{correspondence-guided attention} (CGA). The CGA comprises $B$ transformer-based blocks, each consisting of a standard self-attention module followed by a top-$k$ sparse cross-attention (TKCA) module. %The CGA effectively exploits the interconnections established through correspondence learning in VLFA, allowing our network to effectively link local regions of the query with those in the multi-view images. 
% \CJ{Check: legend font color.}
}
\label{fig:cgsa}
 \vspace{-0.4cm}
\end{figure}

\subsection{Correspondence Matching Transformer}
\label{sec:cmt}
CMT uses ResNet18 feature pyramid network~\cite{Lin_2017_CVPR} as the feature encoder, which is denoted as $\varphi:\mathbb{R}^{3 \times H \times W} \rightarrow \mathbb{R}^{d \times h \times w} $ where the input is down-scaled 8 times through the network ($h=H/8$ and $w=W/8$).
Once we extract the features of $\bq$ and for each $\bv$,  $\varphi(\bq)$ and $\varphi(\bv)$ respectively, we reshape each of them to be $d \times n^q$ dimensional matrices, denote them as $\bff^q$ and $\bff^{v}$ respectively, where $n^q=h\times w$.
Each column in $\bff^q$ and $\bff^{v}$ corresponds to a $d$ dimensional local feature.
We use $\bff[.j]$ notation to indicate $j$-th local feature or patch encoding, as each encoding approximately corresponds to a local patch in the input image due to locality in the convolutional encoder.
Next, we describe key components of the CMT including the 3DPE and CGA modules.

\paragraph{3D Positional Encoding (3DPE).}
% \label{sec:3dpe}
%While the features of the rendered multi-view images obtained using the CNN encoder contain valuable visual information about the reference shape, it is also important to extract 3D positional details to fully leverage the reference 3D shape. 
While the multi-view representation allows for a simple and efficient model design through a shared feature encoder for our task, it also makes 3D information less accessible and hence hampers relating local features across different views accurately.
To mitigate this problem, we propose complementing the multi-view images with 3D information.
For each patch encoding $\bff^v[.j]$, we first locate the corresponding image patch in $\bv$ and then compute the 3D position of the corresponding patch $\bx_j \in \mathbb{R}^3$ in the world coordinates 3D {using the known camera parameters and depth maps.}
Then we use Fourier encoding to obtain a higher-dimensional vector for each $\bx_j$ and further process it through an MLP block to obtain a $d$ dimensional 3DPE.
Formally, we denote the joint mapping by $\gamma: \mathbb{R}^{3} \rightarrow \mathbb{R}^{d}$.

Compared to the 2D standard positional encoding used in transformer models~\cite{dosovitskiy2020image}, 3DPE encodes 3D object geometry in the world space.
For each $\bff^{v}$ including $n^q$ patch encodings, we compute a corresponding $d \times n^q$ dimensional matrix $\bp^{v}$.
For the next steps, we gather $\bff^{v}$ and $\bp^{v}$ over $N$ views, and concatenate each set along their second dimensions, resulting in $\bF^{v} \in \mathbb{R}^{d \times n^v}$ and $\bP^{v} \in \mathbb{R}^{d \times n^v}$ respectively where $n^v = N \times n^q$.
Augmenting $\bF^{v}$ with $\bP^{v}$ results in a novel hybrid 2D-3D representation by incorporating explicit 3D information into the 2D multi-view images.

% We introduce a 3D positional encoding module, denoted as 3DPE, which generates 3D positional features corresponding to each local patch in the multi-view images. We argue that the CMT model can leverage these 3D priors to make better predictions instead of focusing on each view in isolation. 
% At first, we calculate 3D world coordinates for individual local patches within the multi-view images. We accomplish this through inverse projection, which maps 2D points in an $h \times w$ feature map to their corresponding 3D points by utilizing the known camera intrinsic and extrinsic parameters. The obtained 3D coordinates are normalized to lie in $[-1,1]$. Instead of using the raw 3D coordinates directly, we employ Fourier encoding to obtain a higher-dimensional vector for each point and then pass it through an MLP block (both jointly denoted as $F_\gamma$). In the end, the 3DPE outputs 3D positional features $\mathbf{F}_{pos} \in \mathbb{R}^{n^v \times 128}$. Contrary to the typical positional encoding employed in transformer models~\cite{dosovitskiy2020image}, our 3DPE technique fully leverages the 3D geometry to encode actual 3D positions in the world space. 

\paragraph{Correspondence-Guided Attention (CGA).}
% we concatenate these encodings over all the views to obtain a $d\times n^v $ dimensional matrix $\bF_{\text{pos}}$, and further concatenate $\bF^v$ and $\bF_{\text{pos}}$ along their first dimension to obtain a $2d \times n^v $ dimensional matrix $\bar{\bF}^v$.

\begin{figure}[t!]
\begin{center}
   \includegraphics[width=1\linewidth]{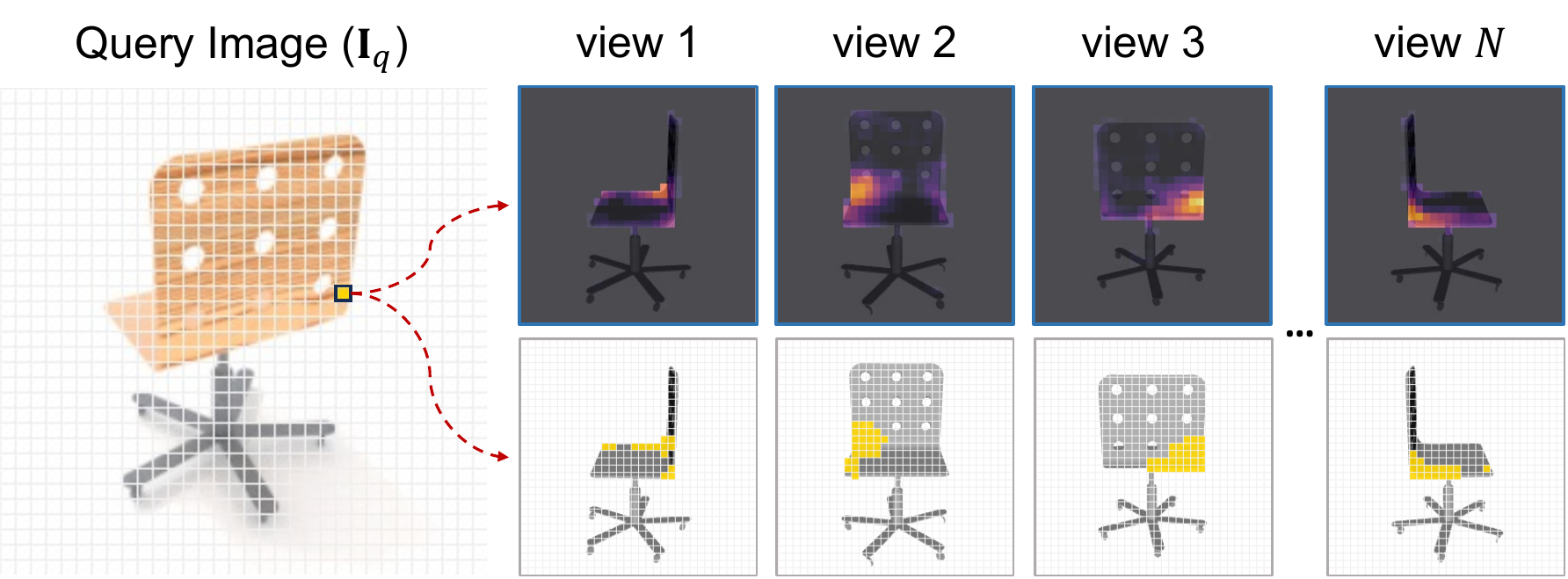}
\end{center}
\vspace{-0.5cm}
\caption{\textbf{Top-$k$ sparse attention-span visualization.} For the query point (yellow), similarity heatmaps (first row) and top-$k$ attention-span (second row) across multiple views are shown.}
\label{fig:topk}
 \vspace{-0.4cm}
\end{figure}

% \label{sec:cga}
The CGA network $\phi$, as illustrated in \cref{fig:cgsa}, takes in $\bff^q$, $\bF^{v}$, $\bP^{v}$ and predicts the anomaly label while efficiently computing the correlations across two modalities.
CGA comprises $B$ consecutive transformer blocks where each block contains multiple operations and is indexed by subscript $b$.
In particular, the block $b$ starts with concatenating $\bF^{v}$ and $\bP^{v}$ along their first dimension, then the resulting  $2d\times n^v$ dimensional matrix is reduced to $d \times n^v$ dimensional $\bar{\bF}^{v}$ matrix through a linear projection layer $\alpha^{(b)}: \mathbb{R}^{2d}\rightarrow \mathbb{R}^{d}$ (\cref{eq:barfv}).
After self-attention operation (\texttt{SA}) is applied to the query features $\bff^{q}_{(b)}$ where $\bff^{q}_{(1)}= \bff^{q}$ (\cref{eq:fqb}), it computes the query $\bm{Q}_{(b)} \in \mathbb{R}^{d \times n^q}$ (\cref{eq:Q}) and key-value matrices $\bm{K}_{(b)}\in \mathbb{R}^{d \times n^v}$, $\bm{V}_{(b)}\in \mathbb{R}^{d \times n^v}$ (\cref{eq:KV}) by applying the linear projections $\bm{W}^{Q,K,V}\in \mathbb{R}^{d \times d}$ respectively.
% a self-attention layer ($\texttt{SA}$) applied to the query features $\bff^{q}$, a linear projection layer $\alpha^{(b)}: \mathbb{R}^{2d}\rightarrow \mathbb{R}^{d}$ at block $b$ to reduce the dimensionality of $\bar{\bF^v}$, a novel top-\textit{k} sparse cross-attention (TKCA) mechanism, and the other standard transformer layers including residual addition, layer normalization, and feed-forward network.
\vspace{-.2cm}
\begin{flalign}
   &\bar{\bF}^{v}_{(b)} \leftarrow \alpha_{(b)}(\begin{bmatrix}  \bF^{v} \\ \bP^{v} \end{bmatrix}) \label{eq:barfv}\\
   &\bar{\bff}^q_{(b)}  \leftarrow \texttt{SA}(\bff^{q}_{(b)}) \label{eq:fqb}\\
   & \bm{Q}_{(b)}   \leftarrow \bm{W}^{Q} \bar{\bff}^q_{(b)} \label{eq:Q}\\ 
   & \bm{K}_{(b)}  \leftarrow \bm{W}^{K} \bar{\bF}^{v}_{(b)},\quad
   \bm{V}_{(b)}  \leftarrow \bm{W}^{V} \bar{\bF}^{v}_{(b)} & \label{eq:KV} \\ 
   & \bm{O}_{(b)}  \leftarrow \texttt{TKCA}{(\bm{Q}_{(b)},\bm{K}_{(b)},\bm{V}_{(b)},\bm{M})} \label{eq:tkcaob}\\
   & \bm{O}_{(b)}  \leftarrow \texttt{Norm}{(\bm{O}_{(b)}+\bm{Q}_{(b)})} \label{eq:ob1}\\
   & \bm{O}_{(b)}  \leftarrow \texttt{Norm}{(\texttt{FFN}(\bm{O}_{(b)})+\bm{O}_{(b)})} \label{eq:ob2} \\
   & \bff^q_{(b+1)}  \leftarrow \bm{O}_{(b)} \label{eq:qbplus1}
\end{flalign}
% In particular, at each block $b$, we apply SA to the query features to enhance its features by capturing pertinent long-range contextual information between the patches of the query image.
% Next, we pass the self-attended query $\bq^{(b)}$ \AB{$\bff_q^{(b)}$?} along with the projected the hybrid multi-view features $g_{\alpha^{(b)}}(\bar{\bF^v})$ through the TKCA.
% The CGA network comprises $L$ transformer-based layers, each consisting of a standard self-attention block followed by a top-\textit{k} sparse cross-attention (TKCA) block. At first, we flatten the spatial dimension of the query features $\mathbf{f}_q$ to obtain a sequence of feature maps of size $d \times 128$, where $d = h \times w$. Then $\mathbf{f}_q$ undergoes processing in the self-attention block, enhancing the features by capturing pertinent long-range contextual information within the query image. We concatenate the previously obtained multi-view representation $\mathbf{F}_{v}$ with its corresponding positional features $\mathbf{F}_{pos}$ and then apply a linear projection to obtain $\mathbf{F}$ with dimensions $(N \times d) \times 128$. We pass $\mathbf{F}$ along with the self-attended query representation ${\mathbf{f}_q}'$ through the TKCA block. 
% \noindent \textbf{Top-\textit{k} sparse cross-attention (TKCA).}
Next, we pass $\bm{Q}_{(b)},\bm{K}_{(b)},\bm{V}_{(b)}$ to our top-\textit{k} sparse cross-attention (TKCA) module (see \cref{eq:tkcaob}).
Unlike the vanilla cross-attention module in standard transformers~\cite{vaswani2017attention, dosovitskiy2020image} ingesting all tokens for the attention computation, which is inefficient for our task and may introduce noisy interactions with irrelevant features, potentially degrading performance, TKCA calculates the attention between query and only a small subset of relevant multi-view features using a similarity matrix $\bM$:
\begin{equation}
   \texttt{TKCA}{(\bm{Q},\bm{K},\bm{V},\bM)} = \textup{softmax}\left(\mathcal{T}^{\bm{M}}_{k}(\frac{\bm{Q}\bm{K}}{\sqrt{d}}) \right) \bm{V}
   \label{eq:TKCA}
\end{equation} where $\mathcal{T}^{\bm{M}}_{k}$ is given by: 
\begin{equation}
   \mathcal{T}^{\bm{M}}_{k}(\mathbf{A})[ij] =
   \begin{cases}
       A[ij], & \text{if } \bm{M}[ij] \in \texttt{top}_\textit{k} (\bm{M}[i.]) \\
       -\infty,              & \text{otherwise}
   \end{cases}
\end{equation} where $\texttt{top}_\textit{k} (\bm{M}[i.])$ operation selects the $k$ most similar features from multi-view representation (see \cref{fig:topk}) for $i$-th query feature.
To compute $\bM$, we use an auxiliary function $\beta:\mathbb{R}^{d}\rightarrow \mathbb{R}^{d}$, instantiated as a four layered MLP followed by a final channel-wise normalization, that projects $\bff^q$ and each view in $\bF^v$ to a view-agnostic feature space where images corresponding to same object part in 3D has similar representations regardless of their viewpoint.
To obtain the similarity between the query and multi-view patches, we compute the dot product between their projected features:
\begin{equation}
\bM = \beta(\bff^q)^{\mathrm{T}} \beta (\bF^{v}) \in \mathbb{R}^{n^q \times n^v}.
\end{equation}
In contrast to other transformer architectures using sparse attention \cite{wang2022kvt}, TCKA chooses the top-$k$ elements based on a different source of information, geometric correspondences computed across two modalities, and enables an efficient computation of the cross-attention, as the same $\bM$ is used throughout the transformer blocks.
After the cross-correlation, the standard residual addition, normalization and feedforward (FFN) layers are applied to obtain $\bff^{q}$ as input to the next block $b+1$ (\cref{eq:ob1,eq:ob2,eq:qbplus1}).
Note that we use multiple heads, concatenate the outputs from multi-head attention and then derive the final attention results through linear projection. We append a learnable token denoted as $\texttt{[tok]}$ to construct the query inputs of the CGA network. 
Through the transformer blocks, the output state of the \texttt{[tok]} token develops a consolidated representation enriched by learned shape-image correlation, which is used as input to the classification head. 

% Here, we use matrix multiplication to calculate the similarity between the view-agnostic features obtained via the projection head $R_\phi$. The $\texttt{Top-\textit{k}}$ operation selects the $k$ most similar features from multi-view representation (see Fig.~\ref{fig:topk}) for each query. $\boldsymbol{\mathbbm{1}}[\cdot]$ denotes binary mask formation based on the obtained top-$k$ indices. 

% The keys $\mathbf{K}$ and values $\mathbf{V}$ are derived from $\mathbf{F}$ while queries $\mathbf{Q}$ are obtained from ${\mathbf{f}^q}'$ via linear projections. Guided by the sparse connectivity mask $\mathbf{m}$, we compute only the nonzero elements of the sparse attention matrix. The output of sparse dot-product attention is defined as 
% \begin{equation}
% Attn(\mathbf{Q},\mathbf{K},\mathbf{V}) = \textup{softmax}\left(\mathcal{M}_\mathbf{m}\circ\frac{\mathbf{Q}\mathbf{K}}{\sqrt{d_k}}\right)\mathbf{V}
% \end{equation}
% where $\mathcal{M}_\mathbf{m}$ denotes a masked operation 
% \begin{equation}
% \left[\mathcal{M}_\mathbf{m}\circ\mathbf{A})\right]_{ij} =
% \begin{cases}
%     A_{ij}, & \text{if } m_{ij} = 1\\
%     -\infty,              & \text{otherwise}
% \end{cases}
% \end{equation}
 %computing the sparse attention matrix A costs only as many FLOPs as there are nonzero elements in the con- nectivity mask M. 

\subsection{View-Agnostic Local Feature Alignment}
\label{sec:VLFA}
As discussed above, image-level supervision alone is too weak to capture fine localized correlations between $\bq$ and $\mcV$. 
Thus, we introduce an auxiliary task, VLFA that aims to densely align corresponding parts between query images and related views.
Through $\beta$, we learn to map $\bff^q$ and $\bff^v$ to a view-agnostic space such that their local features corresponding to the same object part are mapped to a similar point regardless of the viewpoint from which the image is captured.
As the viewpoint of $\bq$ is unknown, the ground-truth correspondences between query and reference views cannot be obtained through inverse rendering. 
% To address this challenge, we propose two strategies.

To this end, we use a self-labeling strategy to generate pseudo-correspondences by finding the most similar local feature in the reference view to each local feature in the query at each training step, after mapping their features to the view-invariant space and normalizing them:
\begin{equation}
   \hat{c}_i = \underset{j}{\textup{argmax}}\; {\bz^{q}_i}^\mathrm{T} \bz^{v}_j,
   \label{eq:pseudo}
\end{equation} where $\bz^{q}_i= \frac{\beta(\bff^{q}[.i])}{\Vert \beta(\bff^{q}[.i])\Vert}$ and $\bz^{v}_j= \frac{\beta(\bff^{v}[.j])}{\Vert \beta(\bff^{v}[.j])\Vert}$.
We compute the pseudo-label for each $\bz^{q}_i$ and store them in a look-up table $\hat{\mathcal{P}}(\bq,\bv,i)=\hat{c}_i$.
In another one $\hat{\mathcal{N}}(\bq,\bv,i)$, we store the remaining set of reference view and index values that are not the corresponding location.
Then, using $\hat{\mathcal{P}},\hat{\mathcal{N}}$ as positive and negative correspondences respectively, we minimize a contrastive loss $\ell_\textup{va}(\bq,\bv)$ over each $\bq$-$\bv$ pair:
\begin{equation}
   % \ell_\textup{va}(\bq) =  
   % \frac{1}{n^q \cdot N}
   \sum_{i=1}^{n^q}
   -\textup{log}\frac{\textup{exp}({\bz^{q}_{i}}^{\mathrm{T}} \bz^{v}_{+} /\tau)}{\textup{exp}({\bz^{q}_{i}}^{\mathrm{T}} \bz^{v}_{+}/\tau) + \underset{{j \in \hat{\mathcal{N}}(\bq,\bv,i)}}{\sum}  \textup{exp}({\bz^{q}_{i}}^{\mathrm{T}} \bz^{v}_{j}/\tau)},
   \label{eq:contrastive}
\end{equation} where $\tau$ is a temperature parameter and $\bz^{v}_{+}=\bz^{v}_{\hat{\mathcal{P}}(\bq,\bv,i)}$.
Due to the cost of computing the pseudo-correspondences for all query features across all views, we compute them only for a random subset of query features across randomly sampled views at each training iteration.

% However, learning correspondences between two modalities is challenging due to the missing groundtruth labels and also the domain gap between them.

Learning correspondences through self-learning alone in the presence of the domain gap between query and reference views is a noisy process.
Hence, we also exploit the known viewpoints of the multi-view images by densely aligning their local features in each view pair $\bv,\bv'$ after computing the ground truth dense correspondences between them 
and discarding the ones that are occluded in one of the views.
The key assumption here is that aligning different views by using their ground truth labels enables a more accurate correspondence learning between query images and views, as the parameters of the projection $\beta$ are shared across two domains.
As before, we form two look-up tables $\mathcal{P}(\bv,\bv')$ and $\mathcal{N}(\bv,\bv')$ to store the positive and negative correspondences between two views, and randomly subsample them. % for computational reasons in practice.
After mapping them to the view-invariant space and normalizing them, we compute and minimize \cref{eq:contrastive} for the pairs in the look-up tables.

\begin{figure*}[t!]
\begin{center}
   \includegraphics[width=1\linewidth]{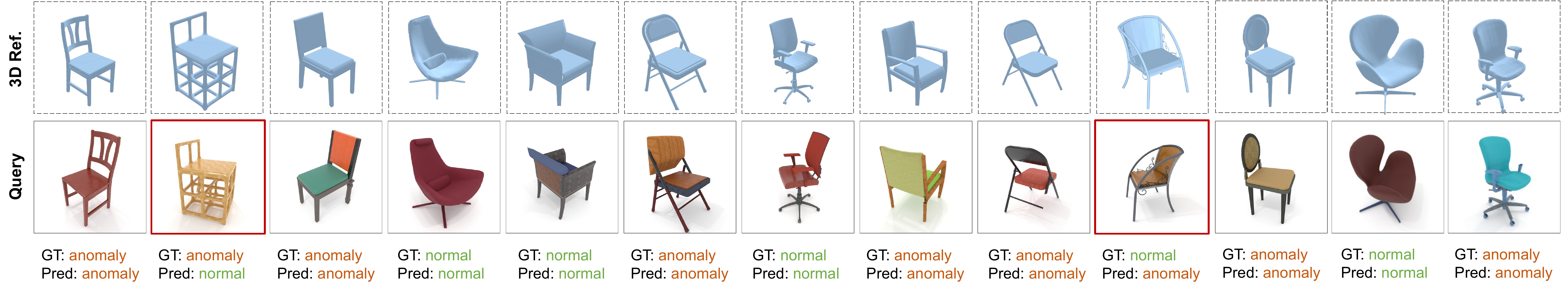}\vspace{-0.5cm}
\end{center}
\vspace{-2mm}
\caption{Anomaly detection results on the test set using our proposed CMT framework. Incorrect predictions are marked in red.}
\label{fig:visres}
 \vspace{-0.4cm}
\end{figure*} 

The objective in \cref{eq:optceonly} can be rewritten as:
\begin{equation}
 \mathcal{L}_{\text{bce}}(\mathcal{D})+ a \mathcal{L}_\textup{va}(\hat{\mathcal{P}},\hat{\mathcal{N}}) + (1-a) \mathcal{L}_\textup{va} (\mathcal{P},\mathcal{N}) 
   \label{eq:optfinal}
\end{equation} where $\mathcal{L}_\textup{va}(\hat{\mathcal{P}},\hat{\mathcal{N}})$ and $\mathcal{L}_\textup{va} (\mathcal{P},\mathcal{N})$ are the alignment loss functions over query-view pairs and view-view pairs respectively, $a$ is a loss balancing weight  set to $0.5$.

\section{Experiments} 
\label{sec:exp}

\noindent \textbf{Implementation Details}: The encoder $\varphi$ takes a $3 \times 256 \times 256$ image as input and returns a $128 \times 32 \times 32$ feature block. %$\gamma$ consists of a Fourier encoding layer followed by an MLP block with two linear layers. %$R_\phi$ is based on $1 \times 1$ convolution projection layers. 
Within the CGA network, we employ three transformer blocks ($B=3$), and each applies 8-headed attention. %Following common practice~\cite{dosovitskiy2020image}, we incorporate fixed sinusoidal positional encodings into the query features before forwarding them to the attention network. In the case of multi-view representation though, 3DPE already encodes positional information, so we don't add any further positional encoding for the multi-view representation. 
The value of $k$ in TCKA is set to $100$. 
During training, we randomly select a subset of $N=10$ views, and during testing, we utilize all 20 views. 
We apply basic data augmentation to the query images, which includes random horizontal flips and random cropping of $224 \times 224$ regions, followed by resizing the cropped regions back to the original size of $256 \times 256$. 
We train our model for 20 epochs using 4 Titan RTX GPUs, maintaining a batch size of 8 in each GPU, and use the Adam optimizer with a learning rate of $2\times10^{-5}$. 
% We refer to the supplementary material for more implementation details.
We refer to the supplementary for more details.
% Our dataset, codes, and models will be made public. 
%\CJ{Can move the whole paragraph to the suppl.}

\subsection{Results}
Since there is no related public benchmark for our task, we define several challenging baselines to evaluate our CMT.
We report the quantitative results using two evaluation metrics -- the area under the ROC curve (AUC) and accuracy in \cref{tab:main_table}, and provide qualitative results in \cref{fig:visres}. 

%\CJ{Where is the definition of metrics?}

\noindent \textbf{Importance of 3D reference
shape.} %A chair instance could be considered as an anomaly solely in relation to a predefined reference shape, while independently it could appear to be a normal or regular chair. This leads to ambiguity if we don't provide any reference to the model. Furthermore, given the complexity of our dataset, several instances (\ie, a slightly tilted leg) necessitate a reference shape for accurate anomaly classification.  
To assess the significance of using the reference shape, we establish baselines that solely rely on the query image for detecting anomalies. 
As our first baseline, we use a ResNet18-FPN model that takes in only query images as input.
For the next two baselines, we add three self-attention blocks to ResNet18-FPN and use a ViT~\cite{dosovitskiy2020image} respectively.
\cref{tab:main_table} shows that the reference 3D shape is crucial to good performance while CMT outperforms the baselines by more than 10\% in accuracy.

\begin{table}
\begin{center}
\caption{Quantitative Comparison of the proposed CMT framework with several baselines in terms of area under the ROC curve (AUC) and accuracy score. For both metrics, higher is better. 
% \CJ{Check: to be filled?}
}
\vspace{-2mm}
\label{tab:main_table}
\setlength{\tabcolsep}{8pt}
\scalebox{0.70}{
\begin{tabular}{c|l|c|c}
\toprule[0.4mm]
\rowcolor{mygray} 
\cellcolor{mygray} 3D Ref. &Methods &
  AUC ($\uparrow$) &  Accuracy ($\uparrow$) \\ \midrule
\multirow{3}{*}\xmark &  ResNet18-FPN~\cite{Lin_2017_CVPR} &  74.6  &  64.7  \\
 & ResNet18-FPN w/ SA blocks  &  75.2  & 65.1 \\
 & Vision Transformer (ViT) ~\cite{dosovitskiy2020image}  & 75.4  & 65.2 \\
  \midrule
  \multirow{8}{*}\cmark 
  & LFD [CVPR'19]~\cite{grabner2019location}  & - & 64.9\\
  & Lin~\etal [ICCV'21]~\cite{lin2021single}  & - & 67.8 \\  
  \cmidrule{2-4}
 & \texttt{\textit{A}}: CMT \textit{(w/o CGA, VLFA, 3DPE)} &  76.1
  & 66.8 \\
 & \texttt{\textit{B}}: CMT \textit{(w/o VLFA, 3DPE)} &   76.3 &  67.1 \\
 & \texttt{\textit{C}}: CMT \textit{(w/o CGA, 3DPE)}  &   80.8 &  72.3 \\
 & \texttt{\textit{D}}: CMT \textit{(w/o 3DPE)}  &  82.6  &  73.7  \\  \cmidrule{2-4}

 &  \textbf{Ours: CMT}   & \textbf{84.7}  &  \textbf{75.4} \\
\bottomrule[0.4mm]
\end{tabular}
}
\end{center}
\vspace{-0.5cm}
\end{table}

\begin{table}
\begin{center}
\caption{ Ablation of view-agnostic alignment loss. The loss function comprises two components: $\mathcal{L}_\textup{va} (\mathcal{P},\mathcal{N})$ and $\mathcal{L}_\textup{va}(\hat{\mathcal{P}},\hat{\mathcal{N}})$. Optimal performance is achieved when both losses are combined.
}
\vspace{-2mm}
\label{tab:ablation_loss}
\setlength{\tabcolsep}{12pt}
\scalebox{0.7}{
\begin{tabular}{c|c|c|c}
\toprule[0.4mm]
\rowcolor{mygray} 
\cellcolor{mygray}$\mathcal{L}_\textup{va} (\mathcal{P},\mathcal{N})$  & $\mathcal{L}_\textup{va}(\hat{\mathcal{P}},\hat{\mathcal{N}})$  &
  AUC ($\uparrow$) &  Accuracy ($\uparrow$) \\ \midrule
    \xmark & \xmark  & 78.5  &  68.3 \\
     \xmark & \cmark &  78.6 & 68.1  \\
    \cmark & \xmark & 81.6 
  & 73.7 \\
    \cmark & \cmark &  \textbf{84.7} &   \textbf{75.4} \\

\bottomrule[0.4mm]
\end{tabular}
}
\end{center}
\vspace{-0.8cm}
\end{table}

\noindent \textbf{Comparison with related work}.
As there is no prior work designed for our problem, we take two recent image-based 3D shape retrieval techniques~\cite{grabner2019location, lin2021single} that learn to embed 2D images and 3D shapes into a common feature space and perform metric learning using a triplet loss.
Once we train them in our dataset, we evaluate them by using the distance between the query and reference shape embeddings to obtain the classification score after a thresholding step.
% We report the accuracy score in comparison with our CMT in \cref{tab:main_table}.
Based on results in \cref{tab:main_table}, we argue that these methods fail to locate subtle variations in geometry, as the cross-modal correlations are only learned at the image level missing fine-grained local correspondence learning.

\noindent \textbf{Ablation of CGA, VLFA and 3DPE.} 
% Here we study the impact of three components, VLFA, CGA, and 3DPE on the results.
Our first baseline (\texttt{\textit{A}}) includes none of the three components but a standard cross-attention module to relate two modalities using all local patches.
Surprisingly, \texttt{\textit{A}} obtains only a 1.6\% accuracy improvement over the query-only baseline, indicating its inability to fully leverage the reference shape. 
% Next, we define two additional baselines \texttt{\textit{B}} and \texttt{\textit{C}}. 
Baseline \texttt{\textit{B}} includes only the CGA component with the top-$k$ sparse cross-attention, baseline \texttt{\textit{C}}, contains the VLFA but with a standard cross-attention. 
While baseline \texttt{\textit{B}} does not show much improvement over \texttt{\textit{A}}, baseline \texttt{\textit{C}} performs significantly better than \texttt{\textit{A}} obtaining a 5.2\% accuracy gain.
This clearly demonstrates the importance of the auxiliary task where we learn matching correspondences for the AD task and that the CGA fails to acquire meaningful correspondences in the absence of the VLFA.
Baseline \texttt{\textit{D}} that employs both CGA and VLFA further boosts the performance of \texttt{\textit{C}} through its selective sparse attention mechanism.
Finally, our model, which includes all the components, outperforms \texttt{\textit{D}} thanks to the introduction of 3DPE that facilitates corresponding matching across different views.

% while in baseline \texttt{\textit{C}}, we incorporate the VLFA but utilize a standard cross-attention. 
% While baseline \texttt{\textit{B}} does not show much improvement over \texttt{\textit{A}}, baseline \texttt{\textit{C}} performs significantly better than \texttt{\textit{A}} obtaining an absolute 5.2\% accuracy gain.  In another experiment (\texttt{\textit{D}}), we employ both CGA and VLFA, resulting in a 1.4\% improvement over \texttt{\textit{C}}. Based on the results of the aforementioned experiments (\texttt{\textit{A}}-\texttt{\textit{D}}), it can be concluded that the most substantial improvement occurs when VLFA and CGA are employed simultaneously. In the absence of VLFA (as in \texttt{\textit{B}}), the CGA fails to acquire meaningful correspondences, whereas with only the VLFA (as in \texttt{\textit{C}}), the standard cross-attention can still leverage the learned view-agnostic representation, improving the performance. However, with both VLFA and CGA (as in \texttt{\textit{D}}), the network can concentrate on corresponding regions via a sparse attention mechanism, thereby enhancing the overall performance.  The accuracy is further improved by the introduction of 3DPE (+1.7\%) to our final CMT architecture. 

\noindent \textbf{Ablation of loss functions.} 
\cref{tab:ablation_loss} reports an ablation study on the loss functions used for learning the view-agnostic representation. 
Utilizing only query-view alignment loss ($\mathcal{L}_\textup{va}(\hat{\mathcal{P}},\hat{\mathcal{N}})$) does not yield any advantages (row 2) over not employing any alignment loss (row 1). However, employing the view-view alignment ($\mathcal{L}_\textup{va}(\mathcal{P},\mathcal{N})$) alone leads to improved results (row 3). The optimal result is achieved when both components are combined (row 4). 

\begin{figure}[t!]
\begin{center}
   \includegraphics[width=0.88\linewidth]{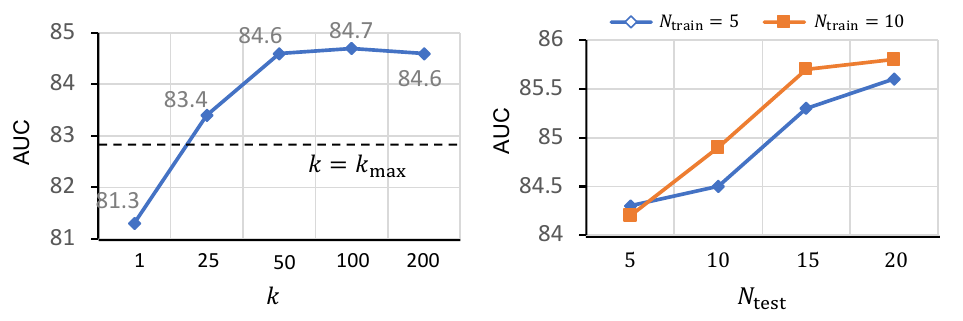}\vspace{-0.2cm}
\end{center}\vspace{-0.5cm}
\caption{(left) AUC score under different values of $k$. (right) Impact of numbers of multi-view images during training and testing. }
\label{fig:ablation_k_N}
 \vspace{-0.2cm}
\end{figure}

\noindent \textbf{Sensitivity to $k$.} 
We analyze the performance under different values of $k$ in \cref{fig:ablation_k_N} (left). 
Compared to the maximum possible $k_{\textup{max}}$ that is $N\times32\times32$, we analyze significantly smaller $k$ values and, among them, show $k = 100$ yields the best result. 
Using all available tokens results in deteriorated performance (shown as the dotted horizontal line) showing that our top-$k$ sparse attention is effective in eliminating the noisy patches by using only the $k$ top-related ones.

\noindent \textbf{Sensitivity to $N$.} 
\cref{fig:ablation_k_N} (right) depicts the analysis on 
the number of input views for training and testing,  $N_{\textup{train}}$ and $N_{\textup{test}}$ respectively. 
To this end, we train two separate CMT models with 5 and 10 views, and evaluate each using 5, 10, 15, and 20 views at test time. 
The plot shows that, while increasing views in both training and testing helps, training with few views and testing on more views can provide a good tradeoff between training time and performance.
% as $N_{\textup{test}}$ increases, the AUC score also improves. 

% \begin{figure}[t!]
% \begin{center}
%    \includegraphics[width=1\linewidth]{figs/bbox.pdf}\vspace{-0.4cm}
% \end{center}\vspace{-0.2cm}
% \caption{Bounding box regression results. We compare our proposed CMT with ViT which does not use the 3D reference. 
% % \CJ{Update: first example in the second row - move the object a bit up.}
% }
% \label{fig:bbox}
%  \vspace{-0.1cm}
% \end{figure}

% \begin{figure}[t!]
% \begin{center}
%    \includegraphics[width=1\linewidth]{figs/pose.pdf}\vspace{-0.4cm}
% \end{center}\vspace{-0.2cm}
% \caption{Query pose estimation using learned correspondence mapping. Failure case(s) are marked in red.}
% \label{fig:pose}
%  \vspace{-0.2cm}
% \end{figure}

\begin{figure}[t!]
\begin{center}
   \includegraphics[width=1\linewidth]{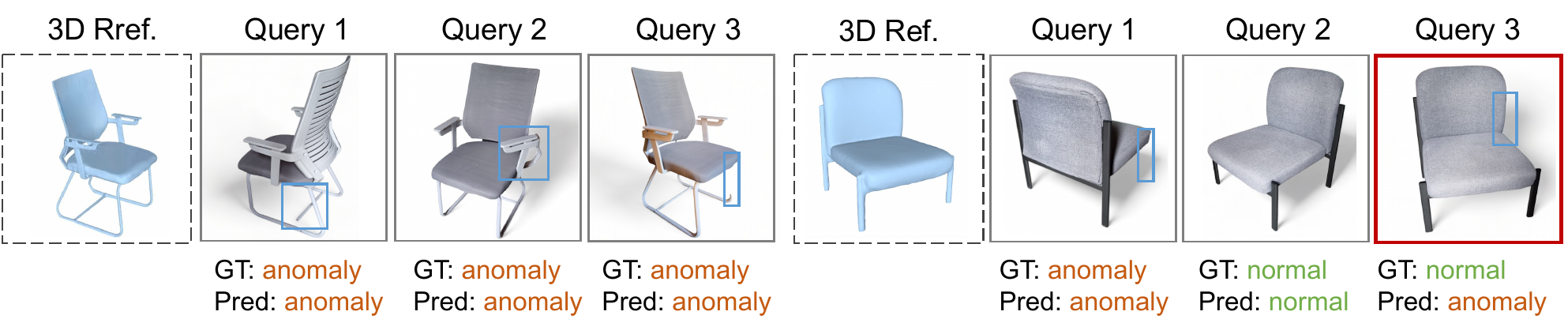}\vspace{-0.4cm}
\end{center}\vspace{-0.2cm}
\caption{Evaluation on real data. The predicted anomalies are shown in the blue bounding boxes. 
% \CJ{Update: first example in the second row - move the object a bit up.}
}
\label{fig:realworld}
 \vspace{-0.5cm}
\end{figure}

\noindent \textbf{Viewpoint prediction.} 
As a side product of establishing the correspondences across the query image and views in our model, we could estimate the camera viewpoint in the query image \wrt a reference shape.
To this end, we compute dense correspondences between the query image and each view image, and then calculate the distance between the pixel coordinates of each point in the query image and their predicted correspondences in the multi-view images, and choose the view with the lowest average distance as the approximate viewpoint.
As a baseline, we train a ResNet with the viewpoint supervision on the normal images only, and evaluate it on the test normal query images.
Our model, trained with no viewpoint supervision, achieves a significantly better accuracy (47\% vs 89\%) when predicting the closest view suggesting that our model implicitly learns to relate the query image with the closest views.
%\CJ{Can move the whole paragraph to the suppl.}

%\Cref{fig:pose} illustrates qualitative results showing the chair image and its matched multi-view image with the closest pose.

\noindent \textbf{Evaluation on real data.} 
Here we apply our model, which is trained on the synthetic \dname dataset, on a small set of real chair samples
that contain multiple pairs of the reference 3D shape, query images containing either normal or irregular instances with broken, removed, or misaligned parts from various viewpoints.
Background pixels in query images are removed by using a segmentation method~\cite{kirillov2023segment} in a preprocessing step and also a synthetic shadow was added to match the training images.
To obtain the reference 3D shapes, we take multiple photos of object instances while walking around them, use the 3D reconstruction software~\cite{tancik2023nerfstudio}, and finally apply Laplacian smoothing to post-process it.
% we scan the respective normal chairs and apply Laplacian smoothing as post-processing. 
\cref{fig:realworld} illustrates the results for two regular reference shapes, each paired with three query images.
In 5 out of 6 cases, our method successfully classifies and localizes the anomalous parts, while in the failure case, it incorrectly relates the self-occluded arm with an anomaly.

% We collected a small set of real-world samples (containing both anomaly and normal along with paired 3D shapes). 
% For this purpose, we utilize actual chairs that are either already broken or have easily disassembled parts to simulate anomalies through the removal or misalignment of components.  
% We capture images of both normal and anomalous chairs from various viewpoints to use as query inputs. 
% Before passing the query to the network, we perform background segmentation. 
% To obtain the reference 3D shape, we scan the respective normal chairs and apply Laplacian smoothing as post-processing. 
% \cref{fig:realworld} demonstrates our model's effectiveness in real-world scenarios. 

\noindent \textbf{Anomaly localization.}
Here we adopt our model to localize anomalies in the form of a bounding box, use a bounding box regression head (a 4-layer MLP), and jointly train it with the other network parameters by using L1 regression and generalized IoU loss~\cite{rezatofighi2019generalized}.
This model achieves 56.5\% average precision on our dataset, outperforms a ViT baseline that is trained only on the query images, and obtains 42.6\%.
Moreover, jointly learning the classification and localization further boosts the classification performance to 85.9 (+1.2) AUC and 77.3\% (+1.9) accuracy.
% not incorporate the 3D reference yields a significantly lower AP of 42.6. Notably, adding the regression branch improves the AUC and classification accuracy to 85.9 (+1.2) and 77.3 (+1.9). 
% \Cref{fig:bbox} illustrates the results of bounding box regression.

% For anomaly localization, we adapt our CMT to regress a bounding box locating any anomalous region. Specifically, we incorporate an extra regression head implemented with a 4-layer MLP that takes the output state of the \texttt{[tok]} token as its input. The output is the 4-dim box coordinates. We use smooth L1 loss and generalized IoU loss~\cite{rezatofighi2019generalized} (GIoU loss) as box regression loss in addition to \cref{eq:optfinal}. 
% Our model achieves an AP of 56.5 on our dataset. In comparison, a ViT-based baseline that does not incorporate the 3D reference yields a significantly lower AP of 42.6. Notably, adding the regression branch improves the AUC and classification accuracy to 85.9 (+1.2) and 77.3 (+1.9). Fig.~\ref{fig:bbox} illustrates the results of bounding box regression.

\noindent \textbf{User Perceptual Study.} 
We also evaluated human performance in our task and conducted a study with 100 participants. 
We presented each participant with 10 pairs of reference shapes and query images, with each pair randomly selected from a random subset of 200. 
We observe a human accuracy of 70.6\%, showing that the proposed task is challenging, while our CMT obtains a superior accuracy of 74.8\% on the same subset.
% On the same subset, the CMT obtains a better accuracy of 74.8\%, indicating the effectiveness of our proposed method. 

\section{Conclusion}
In this paper, we have introduced a new AD task, a new benchmark, and a customized solution inspired by quality control and inspection scenarios in manufacturing.
We showed that an accurate detection of fine-grained anomalies in geometry requires a careful study of both modalities jointly.
Our method achieves this goal by learning dense correspondence across those modalities from limited supervision.
Our benchmark and method also have limitations.
Due to the difficulty and cost of obtaining real damaged objects, our dataset contains only shapes and images of synthetic objects and currently is limited to a single yet very diverse category of `chair', presence of only one anomaly in each query image,  focusing only on shape anomalies excluding the appearance based ones such as fading, discolor, texture anomaly.
Moreover, our model assumes that object instances are rigid, and cannot deal with articulations and deformations, and requires an accurate reference 3D shape for accurate detection.

{
    \small
    \bibliographystyle{ieeenat_fullname}
    \bibliography{main}

\begin{thebibliography}{38}
\providecommand{\natexlab}[1]{#1}
\providecommand{\url}[1]{\texttt{#1}}
\expandafter\ifx\csname urlstyle\endcsname\relax
  \providecommand{\doi}[1]{doi: #1}\else
  \providecommand{\doi}{doi: \begingroup \urlstyle{rm}\Url}\fi

\bibitem[Ahmed and Courville(2020)]{ahmed2020detecting}
Faruk Ahmed and Aaron Courville.
\newblock Detecting semantic anomalies.
\newblock In \emph{AAAI}, 2020.

\bibitem[Aubry et~al.(2014)Aubry, Maturana, Efros, Russell, and
  Sivic]{aubry2014seeing}
Mathieu Aubry, Daniel Maturana, Alexei~A Efros, Bryan~C Russell, and Josef
  Sivic.
\newblock Seeing 3d chairs: exemplar part-based 2d-3d alignment using a large
  dataset of cad models.
\newblock In \emph{CVPR}, 2014.

\bibitem[Bergmann et~al.(2019)Bergmann, Fauser, Sattlegger, and
  Steger]{bergmann2019mvtec}
Paul Bergmann, Michael Fauser, David Sattlegger, and Carsten Steger.
\newblock Mvtec ad--a comprehensive real-world dataset for unsupervised anomaly
  detection.
\newblock In \emph{CVPR}, 2019.

\bibitem[Bergmann et~al.(2021)Bergmann, Batzner, Fauser, Sattlegger, and
  Steger]{bergmann2021mvtec}
Paul Bergmann, Kilian Batzner, Michael Fauser, David Sattlegger, and Carsten
  Steger.
\newblock The mvtec anomaly detection dataset: a comprehensive real-world
  dataset for unsupervised anomaly detection.
\newblock \emph{IJCV}, 2021.

\bibitem[Blum et~al.(2021)Blum, Sarlin, Nieto, Siegwart, and
  Cadena]{blum2021fishyscapes}
Hermann Blum, Paul-Edouard Sarlin, Juan Nieto, Roland Siegwart, and Cesar
  Cadena.
\newblock The fishyscapes benchmark: Measuring blind spots in semantic
  segmentation.
\newblock \emph{IJCV}, 2021.

\bibitem[Carrera et~al.(2016)Carrera, Manganini, Boracchi, and
  Lanzarone]{carrera2016defect}
Diego Carrera, Fabio Manganini, Giacomo Boracchi, and Ettore Lanzarone.
\newblock Defect detection in sem images of nanofibrous materials.
\newblock \emph{IEEE Transactions on Industrial Informatics}, 2016.

\bibitem[Chalapathy et~al.(2018)Chalapathy, Menon, and
  Chawla]{chalapathy2018anomaly}
Raghavendra Chalapathy, Aditya~Krishna Menon, and Sanjay Chawla.
\newblock Anomaly detection using one-class neural networks.
\newblock \emph{arXiv preprint arXiv:1802.06360}, 2018.

\bibitem[Chan et~al.(2021)Chan, Lis, Uhlemeyer, Blum, Honari, Siegwart, Fua,
  Salzmann, and Rottmann]{chan2021segmentmeifyoucan}
Robin Chan, Krzysztof Lis, Svenja Uhlemeyer, Hermann Blum, Sina Honari, Roland
  Siegwart, Pascal Fua, Mathieu Salzmann, and Matthias Rottmann.
\newblock Segmentmeifyoucan: A benchmark for anomaly segmentation.
\newblock \emph{arXiv preprint arXiv:2104.14812}, 2021.

\bibitem[Chandola et~al.(2009)Chandola, Banerjee, and
  Kumar]{chandola2009anomaly}
Varun Chandola, Arindam Banerjee, and Vipin Kumar.
\newblock Anomaly detection: A survey.
\newblock \emph{ACM computing surveys}, 2009.

\bibitem[Deecke et~al.(2021)Deecke, Ruff, Vandermeulen, and
  Bilen]{deecke2021transfer}
Lucas Deecke, Lukas Ruff, Robert~A Vandermeulen, and Hakan Bilen.
\newblock Transfer-based semantic anomaly detection.
\newblock In \emph{ICML}, 2021.

\bibitem[Dosovitskiy et~al.(2021)Dosovitskiy, Beyer, Kolesnikov, Weissenborn,
  Zhai, Unterthiner, Dehghani, Minderer, Heigold, Gelly,
  et~al.]{dosovitskiy2020image}
Alexey Dosovitskiy, Lucas Beyer, Alexander Kolesnikov, Dirk Weissenborn,
  Xiaohua Zhai, Thomas Unterthiner, Mostafa Dehghani, Matthias Minderer, Georg
  Heigold, Sylvain Gelly, et~al.
\newblock An image is worth 16x16 words: Transformers for image recognition at
  scale.
\newblock In \emph{ICLR}, 2021.

\bibitem[Feng et~al.(2019)Feng, Hu, Ang, and Lee]{feng20192d3d}
Mengdan Feng, Sixing Hu, Marcelo~H Ang, and Gim~Hee Lee.
\newblock 2d3d-matchnet: Learning to match keypoints across 2d image and 3d
  point cloud.
\newblock In \emph{ICRA}, 2019.

\bibitem[Grabner et~al.(2018)Grabner, Roth, and Lepetit]{grabner20183d}
Alexander Grabner, Peter~M Roth, and Vincent Lepetit.
\newblock 3d pose estimation and 3d model retrieval for objects in the wild.
\newblock In \emph{CVPR}, 2018.

\bibitem[Grabner et~al.(2019)Grabner, Roth, and Lepetit]{grabner2019location}
Alexander Grabner, Peter~M Roth, and Vincent Lepetit.
\newblock Location field descriptors: Single image 3d model retrieval in the
  wild.
\newblock In \emph{3DV}, 2019.

\bibitem[Hejrati and Ramanan(2012)]{hejrati2012analyzing}
Mohsen Hejrati and Deva Ramanan.
\newblock Analyzing 3d objects in cluttered images.
\newblock \emph{NeurIPS}, 2012.

\bibitem[Hendrycks and Gimpel(2017)]{hendrycks2016baseline}
Dan Hendrycks and Kevin Gimpel.
\newblock A baseline for detecting misclassified and out-of-distribution
  examples in neural networks.
\newblock In \emph{ICLR}, 2017.

\bibitem[Kirillov et~al.(2023)Kirillov, Mintun, Ravi, Mao, Rolland, Gustafson,
  Xiao, Whitehead, Berg, Lo, et~al.]{kirillov2023segment}
Alexander Kirillov, Eric Mintun, Nikhila Ravi, Hanzi Mao, Chloe Rolland, Laura
  Gustafson, Tete Xiao, Spencer Whitehead, Alexander~C Berg, Wan-Yen Lo, et~al.
\newblock Segment anything.
\newblock \emph{arXiv preprint arXiv:2304.02643}, 2023.

\bibitem[Krizhevsky and Hinton(2009)]{krizhevsky2009learning}
Alex Krizhevsky and Geoffrey Hinton.
\newblock Learning multiple layers of features from tiny images.
\newblock 2009.

\bibitem[Lamb et~al.(2022)Lamb, Banerjee, and Banerjee]{lamb2022deepjoin}
Nikolas Lamb, Sean Banerjee, and Natasha~Kholgade Banerjee.
\newblock Deepjoin: Learning a joint occupancy, signed distance, and normal
  field function for shape repair.
\newblock In \emph{ACM TOG}, 2022.

\bibitem[LeCun et~al.(1998)LeCun, Bottou, Bengio, and
  Haffner]{lecun1998gradient}
Yann LeCun, L{\'e}on Bottou, Yoshua Bengio, and Patrick Haffner.
\newblock Gradient-based learning applied to document recognition.
\newblock \emph{Proceedings of the IEEE}, 1998.

\bibitem[Li et~al.(2023)Li, Qin, Gao, Yi, Zhu, Guo, and Xu]{li20232d3d}
Minhao Li, Zheng Qin, Zhirui Gao, Renjiao Yi, Chenyang Zhu, Yulan Guo, and Kai
  Xu.
\newblock 2d3d-matr: 2d-3d matching transformer for detection-free registration
  between images and point clouds.
\newblock In \emph{ICCV}, 2023.

\bibitem[Lim et~al.(2013)Lim, Pirsiavash, and Torralba]{lim2013parsing}
Joseph~J Lim, Hamed Pirsiavash, and Antonio Torralba.
\newblock Parsing ikea objects: Fine pose estimation.
\newblock In \emph{ICCV}, 2013.

\bibitem[Lin et~al.(2021)Lin, Yang, Wang, Lai, Jia, Zhao, and
  Gao]{lin2021single}
Ming-Xian Lin, Jie Yang, He Wang, Yu-Kun Lai, Rongfei Jia, Binqiang Zhao, and
  Lin Gao.
\newblock Single image 3d shape retrieval via cross-modal instance and category
  contrastive learning.
\newblock In \emph{ICCV}, 2021.

\bibitem[Lin et~al.(2017)Lin, Dollar, Girshick, He, Hariharan, and
  Belongie]{Lin_2017_CVPR}
Tsung-Yi Lin, Piotr Dollar, Ross Girshick, Kaiming He, Bharath Hariharan, and
  Serge Belongie.
\newblock Feature pyramid networks for object detection.
\newblock In \emph{CVPR}, 2017.

\bibitem[Mo et~al.(2019)Mo, Zhu, Chang, Yi, Tripathi, Guibas, and
  Su]{mo2019partnet}
Kaichun Mo, Shilin Zhu, Angel~X Chang, Li Yi, Subarna Tripathi, Leonidas~J
  Guibas, and Hao Su.
\newblock Partnet: A large-scale benchmark for fine-grained and hierarchical
  part-level 3d object understanding.
\newblock In \emph{CVPR}, 2019.

\bibitem[Pang et~al.(2021)Pang, Shen, Cao, and Hengel]{pang2021deep}
Guansong Pang, Chunhua Shen, Longbing Cao, and Anton Van~Den Hengel.
\newblock Deep learning for anomaly detection: A review.
\newblock \emph{ACM computing surveys}, 2021.

\bibitem[Park et~al.(2018)Park, Rematas, Farhadi, and
  Seitz]{park2018photoshape}
Keunhong Park, Konstantinos Rematas, Ali Farhadi, and Steven~M Seitz.
\newblock Photoshape: Photorealistic materials for large-scale shape
  collections.
\newblock In \emph{SIGGRAPH Asia}, 2018.

\bibitem[Pham et~al.(2020)Pham, Uy, Hua, Nguyen, Roig, and Yeung]{pham2020lcd}
Quang-Hieu Pham, Mikaela~Angelina Uy, Binh-Son Hua, Duc~Thanh Nguyen, Gemma
  Roig, and Sai-Kit Yeung.
\newblock Lcd: Learned cross-domain descriptors for 2d-3d matching.
\newblock In \emph{AAAI}, 2020.

\bibitem[Rezatofighi et~al.(2019)Rezatofighi, Tsoi, Gwak, Sadeghian, Reid, and
  Savarese]{rezatofighi2019generalized}
Hamid Rezatofighi, Nathan Tsoi, JunYoung Gwak, Amir Sadeghian, Ian Reid, and
  Silvio Savarese.
\newblock Generalized intersection over union: A metric and a loss for bounding
  box regression.
\newblock In \emph{CVPR}, 2019.

\bibitem[Ruff et~al.(2018)Ruff, Vandermeulen, Goernitz, Deecke, Siddiqui,
  Binder, M{\"u}ller, and Kloft]{ruff2018deep}
Lukas Ruff, Robert Vandermeulen, Nico Goernitz, Lucas Deecke, Shoaib~Ahmed
  Siddiqui, Alexander Binder, Emmanuel M{\"u}ller, and Marius Kloft.
\newblock Deep one-class classification.
\newblock In \emph{ICML}, 2018.

\bibitem[Saleh et~al.(2013)Saleh, Farhadi, and Elgammal]{saleh2013object}
Babak Saleh, Ali Farhadi, and Ahmed Elgammal.
\newblock Object-centric anomaly detection by attribute-based reasoning.
\newblock In \emph{CVPR}, 2013.

\bibitem[Song et~al.(2007)Song, Wu, Jermaine, and Ranka]{song2007conditional}
Xiuyao Song, Mingxi Wu, Christopher Jermaine, and Sanjay Ranka.
\newblock Conditional anomaly detection.
\newblock \emph{IEEE TKDE}, 2007.

\bibitem[Tancik et~al.(2023)Tancik, Weber, Ng, Li, Yi, Wang, Kristoffersen,
  Austin, Salahi, Ahuja, et~al.]{tancik2023nerfstudio}
Matthew Tancik, Ethan Weber, Evonne Ng, Ruilong Li, Brent Yi, Terrance Wang,
  Alexander Kristoffersen, Jake Austin, Kamyar Salahi, Abhik Ahuja, et~al.
\newblock Nerfstudio: A modular framework for neural radiance field
  development.
\newblock In \emph{ACM SIGGRAPH}, 2023.

\bibitem[Vaswani et~al.(2017)Vaswani, Shazeer, Parmar, Uszkoreit, Jones, Gomez,
  Kaiser, and Polosukhin]{vaswani2017attention}
Ashish Vaswani, Noam Shazeer, Niki Parmar, Jakob Uszkoreit, Llion Jones,
  Aidan~N Gomez, {\L}ukasz Kaiser, and Illia Polosukhin.
\newblock Attention is all you need.
\newblock In \emph{NeurIPS}, 2017.

\bibitem[Wang et~al.(2022)Wang, Wang, Wang, Lin, Chang, Li, and
  Jin]{wang2022kvt}
Pichao Wang, Xue Wang, Fan Wang, Ming Lin, Shuning Chang, Hao Li, and Rong Jin.
\newblock Kvt: k-nn attention for boosting vision transformers.
\newblock In \emph{ECCV}, 2022.

\bibitem[Wang et~al.(2023)Wang, Kannala, and Barath]{wang2023dgc}
Shuzhe Wang, Juho Kannala, and Daniel Barath.
\newblock Dgc-gnn: Descriptor-free geometric-color graph neural network for
  2d-3d matching.
\newblock \emph{arXiv preprint arXiv:2306.12547}, 2023.

\bibitem[Xiao et~al.(2012)Xiao, Russell, and Torralba]{xiao2012localizing}
Jianxiong Xiao, Bryan Russell, and Antonio Torralba.
\newblock Localizing 3d cuboids in single-view images.
\newblock \emph{NeurIPS}, 2012.

\bibitem[Zhou et~al.(2024)Zhou, Li, Jiang, Wang, Zhou, Zhang, and
  Zhao]{zhou2024pad}
Qiang Zhou, Weize Li, Lihan Jiang, Guoliang Wang, Guyue Zhou, Shanghang Zhang,
  and Hao Zhao.
\newblock Pad: A dataset and benchmark for pose-agnostic anomaly detection.
\newblock In \emph{NeurIPS}, 2024.

\end{thebibliography}
}

% WARNING: do not forget to delete the supplementary pages from your submission 
% 
% {
%     \small
%     \bibliographystyle{ieeenat_fullname}
%     \bibliography{main}
% }

\end{document}